\def\year{2019}\relax
\documentclass[letterpaper]{article} 
\usepackage{aaai19}  
\usepackage{times}  
\usepackage{helvet}  
\usepackage{courier}  
\usepackage{url}  
\usepackage{graphicx}  

\usepackage{booktabs}
\usepackage{amsmath,amssymb}
\usepackage{subfigure}
\usepackage{color}
\nocopyright
\begin{document}
%
\title{DropFilter: Dropout for Convolutions}
\author{Zhengsu~Chen \quad Jianwei~Niu	\quad Qi~Tian\\	
}
\maketitle
\begin{abstract}
Using a large number of weights, deep neural networks have achieved remarkable performance on computer vision and natural language processing tasks. However deep neural networks usually use lots of parameters and suffer from overfitting. Dropout is a widely use method to deal with overfitting. Although dropout can significantly regularize fully connected layers in neural networks, it usually leads to suboptimal results when used for convolutional layers. To tackle this problem, we propose DropFilter, a dropout method for convolutional layers. DropFilter considers the filters rather than the neural units as the basic data drop units. Because it is observed that co-adaptations are more likely to occur inter rather than intra filters in convolutional layers. Using DropFilter, we remarkably improve the performance of convolutional networks on CIFAR and ImageNet. 
\end{abstract}

\section{Introduction}
Recently, deep neural networks have enabled breakthroughs on computer vision and natural language processing tasks~\cite{girshick2014rich,long2015fully,graves2013speech,Vinyals2015Show}. Convolutional neural networks (CNNs) are the widely used models on computer vision tasks. Since AlexNet~\cite{krizhevsky2012imagenet} significantly improved the classification performance on ImageNet, many new deep convolutional network architectures (VGG~\cite{simonyan2014very}, Inception~\cite{szegedy2015going,szegedy2016rethinking,szegedy2016inception}, ResNet~\cite{he2016deep}, DenseNet~\cite{huang2016densely} ) are proposed. These networks achieve remarkable success on recognition tasks and usually employ lots of parameters.

 Deep neural networks usually suffer from overfitting. To avoid overfitting, dropout is used in many network architectures~\cite{srivastava2014dropout,simonyan2014very}. Dropout randomly suppresses the outputs of neural units to reduce the co-adaptations between them, which sometimes can significantly improve the networks. The co-adaptive neural units are similar with each other. It is a waste of parameters and may cause overfitting. Dropout is usually used for fully connected layers. When dropout is used for convolutional layers, the network is slightly improved or even gets worse sometimes. Two reasons may account for this problem. First, compared to fully connected layers, convolutional layers use much fewer parameters. Overfitting and co-adaptations are not so serious in convolutional layers. Second, the co-adaptations in fully connected layers are different  from that in convolutional layers.

	Dropout can avoid the co-adaptations between neural units. For fully connected layers, every unit is equivalent in position, which is the main reason why co-adaptations occur. To decrease the co-adaptations, dropout randomly suppresses the units. During the backward pass, the gradients of these units are also suppressed. Consequently, even if the units have similar weights, they will experience different gradients. Their weights will be updated to different directions. The units will become different to each other and co-adaptations will be reduced.
	 
In convolutional networks, the outputs of convolutional layers are a set of feature maps. The outputs in the same position of different feature maps are equivalent in position. However, because of the different receptive field on the input image, the outputs in the same feature map are not equivalent. \textbf{Co-adaptations are more likely to occur between units that are equivalent in position}. Therefore, for convolutional neural networks, the co-adaptations should be reduced inter rather than intra filters. Standard dropout focuses on the co-adaptations between neural units. For the units that are unlikely to produce co-adaptations, using dropout for them may introduce unnecessary noise.

To reduce the co-adaptations between filters, we propose DropFilter in this paper. DropFilter randomly drops some feature maps of convolutional layers. In other words, DropFilter randomly suppresses the outputs of some convolutional filters and that is why it is called DropFilter. Using DropFilter, the filters in convolutional layers will experience different inputs and gradients, which will reduce the co-adaptations between filters.
 
DropFilter is similar to DropPath to some extent. DropPath is a widely used regularization method for multi-path networks~\cite{larsson2016fractalnet,huang2016deep,zoph2017learning}. DropPath randomly drops some paths during training. Every path is a set of feature maps. Thus both DropFilter and DropPath will drop some feature maps during training. DropPath can be seen as a special case of DropFilter. DropPath can only be used for multi-path networks, while DropFilter can be used for all convolutional neural networks.
	
DropFilter and dropout randomly suppress the outputs of convolutional layer. Deep convolutional neural networks usually contain many convolutional layers. If data drop methods are used for every convolutional layer with the same data retaining rate, directly suppressing the outputs is a little coarse for very deep CNNs. The networks may be very sensitive to the data retaining rate and hard to be tuned. To tackle this problem, we propose ScaleFilter. Unlike DropFilter, ScaleFilter scales the feature maps of the outputs with random weights rather than directly sets some of them to zeros. ScaleFilter is similar to DropFilter but is more stable and easier to be tuned.

DropFilter and ScaleFilter are tested on ResNets, WRNs and non-residual networks in this paper. It is observed that DropFilter and ScaleFilter consistently outperform standard dropout. Sometimes standard dropout slightly improve the networks, while DropFilter and ScaleFilter can still bring significant improvement.

\section{Related Work}
\subsection{Dropout} 
Dropout is proposed by~\cite{srivastava2014dropout} to regularize the networks during training. Dropout randomly suppresses the outputs of networks and is widely used for fully connected layers~\cite{krizhevsky2012imagenet,simonyan2014very}. Afterwards,~\citeauthor{Wan2013Regularization} present DropConnect, which randomly suppresses the connections rather than the outputs of neural units. DropConnect performs better in generalization.~\citeauthor{Wang2013Fast} use Gaussian noise for dropout. In their opinions, Gaussian noise is more stable and faster than Bernoulli noise for dropout. As for the dropout retaining rate, instead of using fixed retaining rate, ~\citeauthor{Morerio2017Curriculum} linearly decrease the data retaining rate during training. They propose that co-adaptations will not occur in the beginning. Thus they use big data retaining rate to avoid introducing unnecessary noise in the beginning. In this paper, we first concentrate on the particular co-adaptations in convolutional layers and propose DropFilter and ScaleFilter to regularize CNNs.

\subsection{DropPath}
There are some networks that have many paths~\cite{larsson2016fractalnet,huang2016deep,zoph2017learning}. Residual networks can also be seen as a multi-path networks~\cite{veit2016residual}. DropPath drops some paths to regularize the networks during training, which can only be used for multi-path networks. Stochastic depth ResNets~\cite{huang2016deep} randomly drops the residual block of ResNets. ResNets with stochastic depth is faster during training and outperform constant depth ResNets on some datasets. FractalNet~\cite{larsson2016fractalnet} is a multi-path network which is designed with self-similarity. Using DropPath, FractalNet produces shallow and deep subnetworks that have different speed and accuracy. Thus these subentworks can be used for various tasks. Shake-shake regularization~\cite{gastaldi2017shake} combines the paths in networks with random weights. Based on the their multi-path networks, shake-shake networks achieve state-of-the-art results on CIFAR. The architecture of NASNets~\cite{zoph2017learning} is learned by a reinforcement learning system. NASNets are constructed by multi-path substructures. Using DropPath, the performance of NASNets is significantly improved. Like DropPath, DropFilter randomly drops some feature maps of the networks. However, DropFilter can be used for any convolutional neural network architecture while DropPath can only be used for specially designed multi-path networks.

\section{Methods}
\begin{figure*}[ht]
\begin{center}
\subfigure[]{
\label{fig:path_blocks:a} 
\includegraphics[width=0.25\linewidth]{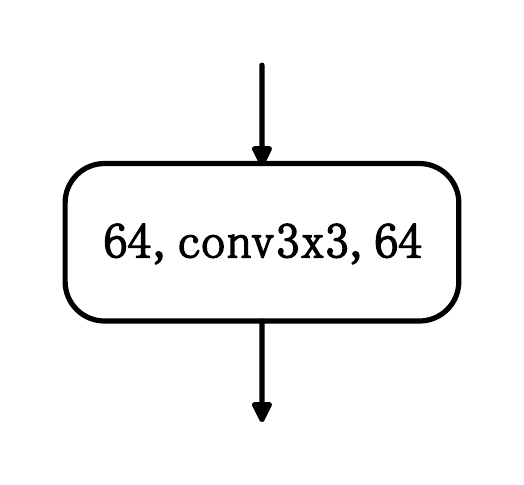}}
\hspace{0in}
\subfigure[]{
\label{fig:path_blocks:b} 
\includegraphics[width=0.5\linewidth]{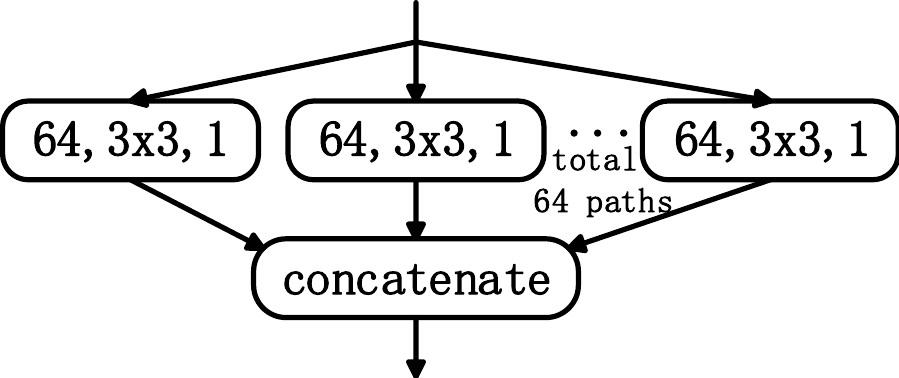}}
\end{center}
\begin{flushleft}
\caption{~Equivalent structures.~(a)~A $3\times3$ convolutional layer. The number of input and output feature map is 64.~(b)~A muti-path convolutional structure. Each path has the same input and produces a feature map. All the output feature maps are concatenated afterwards.}
\end{flushleft}
\label{fig:path_blocks} 
\end{figure*}

\subsection{DropFilter}
Given an input image, $\textbf{x}$ is the input of a convolutional layer for a CNN. $c$ is the filter number of this layer. $Y$ is the output feature maps and $W =\{ \textbf{w}_{i} | 1\leq i \leq c\}$ is the weights of the filters in this layer. We have:

\begin{align}
	\textbf{y}_{i} & = f(\textbf{w}_{i} \textbf{x} + b_{i}) \\
	Y &= \{ \textbf{y}_{i} | 1\leq i \leq c \} \nonumber \label{FilterDrop1}
\end{align}
where $i$ indexes the filters and the output feature maps in the layer. $f$ is the activate function. $b_{i}$ is the bias. Here, the structure of $\textbf{x}$ is reconstructed according to the filters. Consequently the convolution can be presented with multiplication.

Using standard dropout, the outputs will be multiplied by a Bernoulli mask:
\begin{align}
	\textbf{y}^{dropout}_{i} &= \textbf{r}_{i}*\textbf{y}_{i} / p \\
	\textbf{r}_{i} & \sim Bernoulli(p) \nonumber\\ 
	 p       & \in [0,1] \nonumber\\
	 Y_{dropout} &= \{ \textbf{y}^{dropout}_{i} | 1\leq i \leq c\nonumber \}\label{Dropout}
\end{align}
where $p$ is the data retaining rate. The shape of $\textbf{r}_{i}$ is same as $\textbf{y}_{i}$. To compensate the variance shift of dropout, the outputs are scaled by $1/p$ during training~\cite{abadi2016tensorflow}. An alternative choice is removing $1/p$ during training and scaling the outputs by $p$ during testing~\cite{srivastava2014dropout}.   

Using Filter drop, the outputs: 
\begin{align}
	\textbf{y}^{DropFilter}_{i} & =  r_{i}\textbf{y}_{i}/p \\
	r_{i} & \sim Bernoulli(p) \nonumber\\ 
	 p       & \in [0,1] \nonumber\\
	 Y_{DropFilter} &= \{ \textbf{y}^{DropFilter}_{i} | 1\leq i \leq c \}\nonumber\label{FilterDrop2}
\end{align}
where $r_{i}$ is a weight to determinate whether to drop the feature map $\textbf{y}_{i}$. Like standard dropout, DropFilter is also scaled by $1/p$ during training and is removed during testing. 

Dropout aims to avoid the co-adaptations between units. It randomly suppresses the neural units in a convolutional layer equally.
DropFilter focuses on the co-adaptations between filters. It randomly drops whole feature maps to avoid the co-adaptations between filters. Unlike fully connected layers, convolutional layers have more complex structure. The co-adaptations is also different from that in fully connected layers. We believe that this is the main reason why dropout does not work well for convolutional layers. According to structure of convolutions, DropFilter is designed to reduce the co-adaptations in covolutional layers. DropFilter regards the filter rather than the neural unit as the basic co-adaptive unit. This is the main difference between dropout and DropFilter.  

\begin{figure*}[htbp]
\label{fig:diffpq}
\begin{center}
\subfigure[]{
\label{fig:diffpq:a} 
\includegraphics[width=0.45\linewidth]{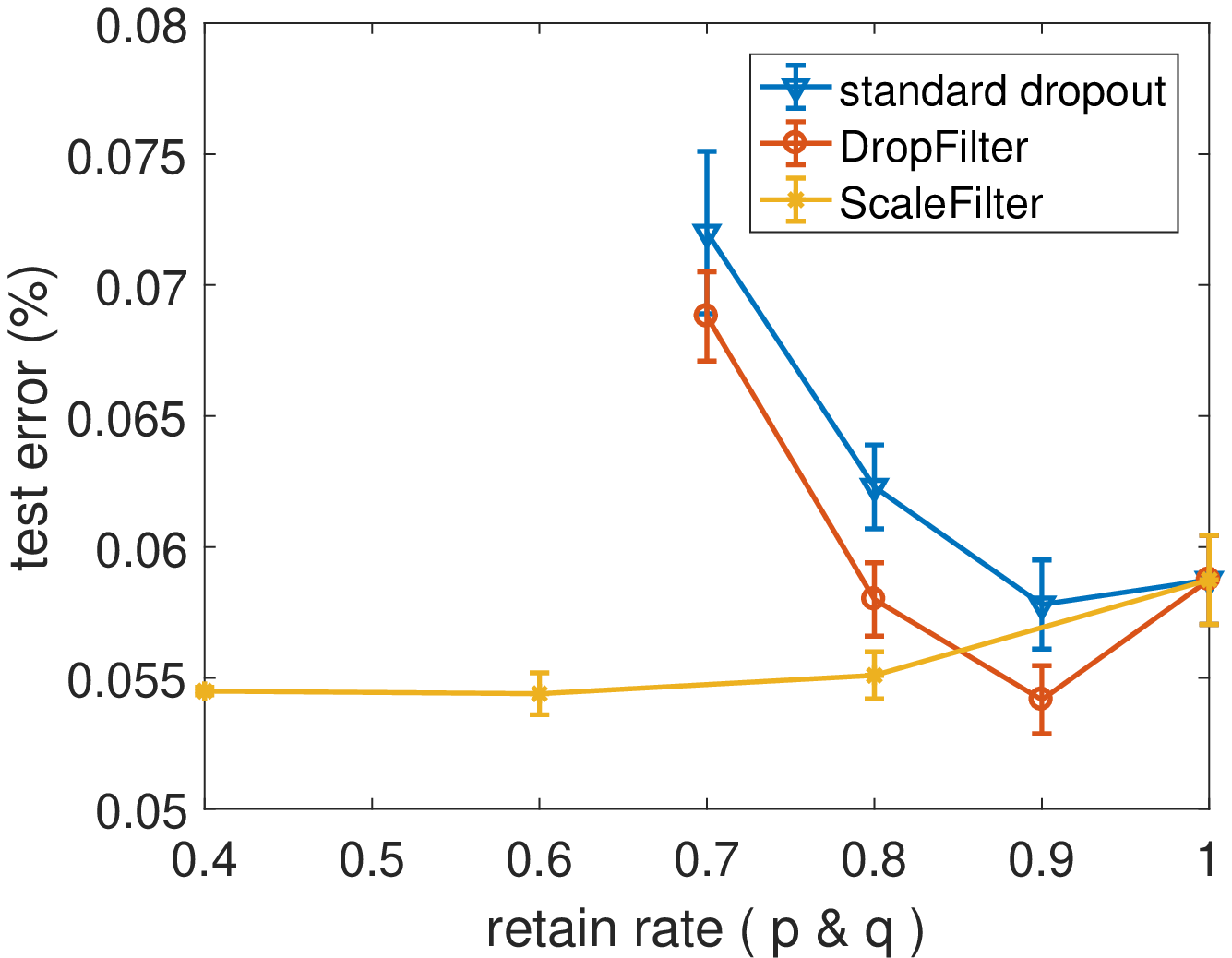}}
\hspace{0in}
\subfigure[]{
\label{fig:diffpq:b} 
\includegraphics[width=0.45\linewidth]{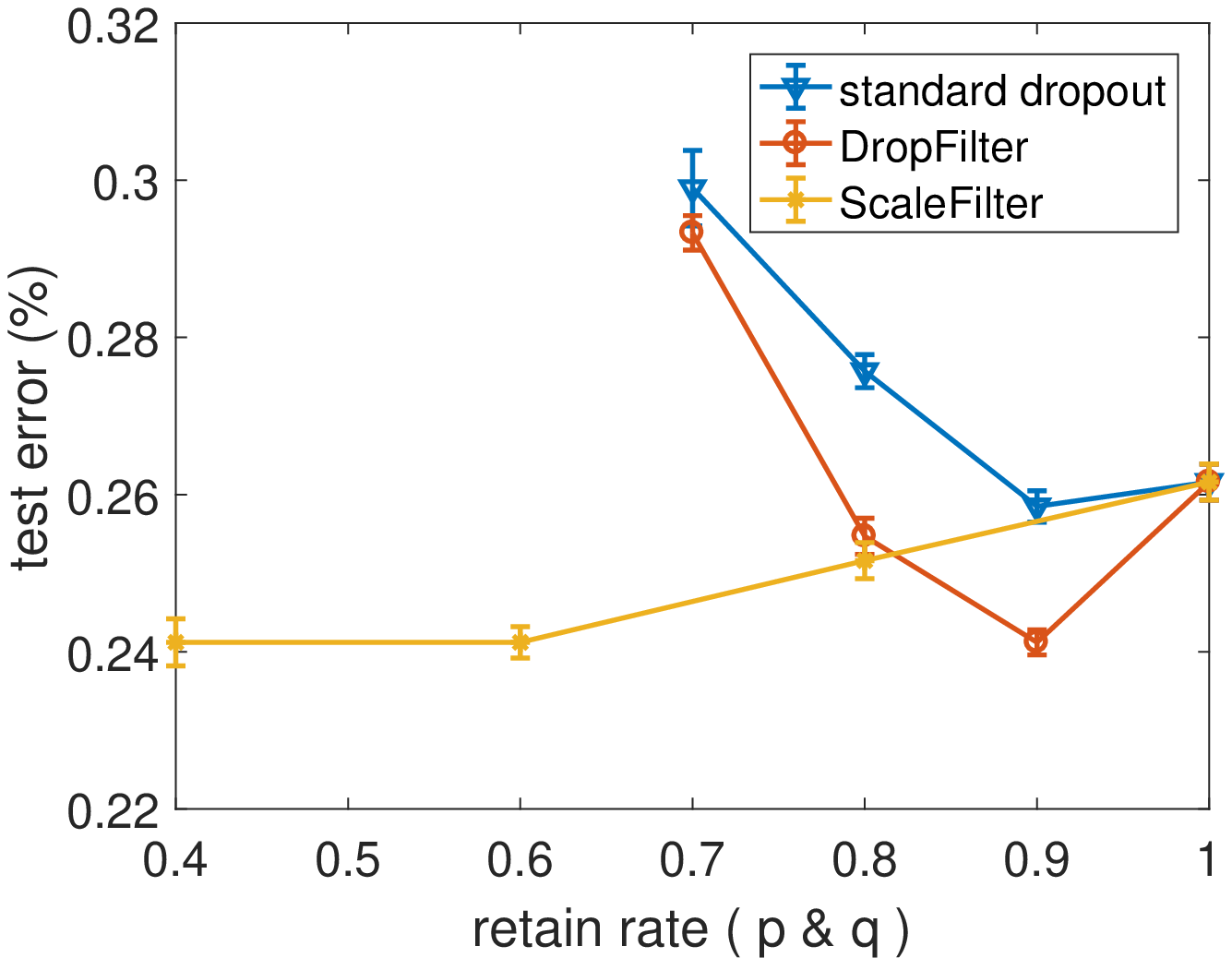}}
\end{center}
\begin{flushleft}
\caption{The test error of different data ratain rate~(~$p$ for standard dropout and DropFilter, $q$ for ScaleFilter).~$p=1$ or $q=1$ means that no data drop method is used. The results are the average of 5 runs.~(a)~The test errors rate of ResNet-110 with different retaining rate on CIFAR-10.~(b)~The test errors of ResNet-110 with different retaining rate on CIFAR-100.}
\end{flushleft}
\label{fig:path_blocks} 
\end{figure*}

\subsection{DropPath and DropFilter}
DropPath is an optional regularization method only for multi-path CNNs. Multi-path networks with DropPath have achieved remarkable results on CIFAR and ImageNet~\cite{zoph2017learning,gastaldi2017shake}. DropPath ramdomly suppresses some paths in multi-path networks. Each path is a set of feature maps. DropPath can be seen as reducing the co-adaptations between paths. DropFilter also drops feature maps. However, DropFilter can be used for all convolutional neural network architecture. In DropFilter, the basic drop unit is the filter. In DropPath, the basic drop unit is the path.

DropFilter is similar to DropPath to same extent. We can understand DropFilter through DropPath.
As shown in Fig.~\ref{fig:path_blocks:a} and Fig.~\ref{fig:path_blocks:b}, each convolutional filter can be considered as a path for the convolutional layer. Afterwards , the outputs of every filter are concatenated as the output of this multi-path structure. When applying DropFilter to a convolutional layer, it is equal to randomly drop some tiny paths for the multi-path structure in Fig.~\ref{fig:path_blocks:b}, which is very similar with DropPath in multi-path networks.

\subsection{ScaleFilter}
DropFilter directly suppresses the outputs of a filter. For some very deep convolutional networks, if DropFilter is applied to every convolutional layer, the network may be very sensitive to the retaining rate $p$. Therefore we propose ScaleFilter, which randomly scales the outputs of the filters rather than directly suppresses them. Using ScaleFilter:
\begin{align}
	\textbf{y}^{ScaleFilter}_{i} & =  r_{i}\textbf{y}_{i} \\
	r_{i} & \sim Uniform(1-q,1+q) \nonumber\\ 
	 q       & \in [0,1]\nonumber\\
	  Y_{ScaleFilter} &= \{ \textbf{y}^{ScaleFilter}_{i} | 1\leq i \leq c \}\nonumber\label{FilterDrop3}
\end{align}
The outputs do not need to be scaled, because the mean of $r_{i}$ is $1$.
During testing, ScaleFilter is removed. 

Both DropFilter and ScaleFilter focus on changing the outputs to reduce the co-adaptations between filters. DropFilter directly suppresses the outputs. ScaleFilter randomly scales the outputs instead. 
ScaleFilter is weaker but more stable than DropFilter. It changes the output data modestly. For some very deep CNNs, the super parameter $q$ for ScaleFilter is easier to tune than $p$ for DropFilter. ScaleFilter sometime can achieve better results.


\section{Experiments}
\subsection{Datasets and Settings}
\subsubsection{CIFAR-10 and CIFAR-100}
CIFAR-10 and CIFAR-100 are colored $32\times32$ natural image datasets that consist of 50,000 images in training set and 10,000 images in testing set. Following the common practice~\cite{he2016deep,zagoruyko2016wide}, these tiny images are padded with four pixels on each side and are randomly cropped with a $32\times32$ window afterwards. Before being fed into the networks, the images are subtracted by channel means and divided by $128$.

\subsubsection{Subset of ImageNet}
DropFilter is further tested on ILSVRC2012 datasets which is a subset of ImageNet~\cite{russakovsky2015imagenet} database. These  dataset contains 1.3M image on training set and 50K images on validation set. ILSVRC2012 is a quite large dataset and requires many resources to run very deep networks upon it. Like miniImageNet~\cite{vinyals2016matching}, we randomly select 100 classes in ILSVRC2012 and randomly select 600 images in each class. Unlike miniImageNet, the images are not resized. During training the images and their horizontal flips are cropped by $224\times224$ window after the short edge being resized to $256$ following~\cite{simonyan2014very}. The trained models are test on the validation set in which the images that do not belong to the selected 100 classes are removed.

\subsection{Implementation Details}
The methods in this paper is test on ResNets~\cite{he2016identity}, WRNs~\cite{zagoruyko2016wide}, ResNeXt~\cite{xie2016aggregated} and plain networks. We use the template in \cite{he2016deep} to construct the ResNets on CIFAR. The ResNets have three stages. Each stage consists of $n$ residual blocks. Each block contains two $3\times3$ layers. The filter numbers in three stage are $\{16,32,64\}$ respectively. We use different $n$ to construct the networks with different depths. The ResNets used in this paper is the  pre-activation ResNets, if not specified.  

Like \cite{zagoruyko2016wide}, WRNs are constructed by widening the ResNets in this paper. WRN-4-20 represents the network with four times more filters in each layer than ResNet-20. 

If not specified, the networks for CIFAR are trained for 200 epochs with 0.1 initial learning rate. The learning rate drops by 0.2 at 60,120 and 160 epochs following \cite{zagoruyko2016wide}. The batch size is 128 and the weight decay is 0.0005. The data drop method is applied to every convolutional layer for ResNets and WRNs. For ResNeXt, data drop is used before and after the $3\times3$ group convolutions. 

On ImageNet, the batch size is 256 and the weight decay is 0.0005. The model is trained for 110 epochs with 0.1 initial learning rate. The learning rate drops by 0.1 at 30, 60, 85, 95 and 105 epochs according to~\cite{wu2016tensorpack}. Data drop regularization is used before and after the $3\times3$ convolutions in bottleneck residual blocks.

All the networks are implemented by tensorflow~\cite{abadi2016tensorflow} and tensorpack~\cite{wu2016tensorpack}. The codes will be released.
\begin{table*}[htp]
  \caption{Test error of different CNNs on CIFAR-10. The data retaining rate (~$p$~) of standard dropout and DropFilter is 0.9 for all tested networks. For ScaleFilter, the retaining rate ($q$) is 0.4 except for ResNet-56 ( 0.6 ). The results are the average of 5 runs.}
  \label{resnet10}
  \centering
  \begin{tabular}{lcccc}
    \toprule
    Network     & without data drop(\%) & standard dropout(\%) & DropFilter(\%)  & ScaleFilter(\%) \\
    \midrule
    ResNet-56& 6.20& 6.08 & {\color{blue}6.07} & {\color{red} 5.65}\\
    \midrule
     ResNet-110& 5.88& 5.78 & {\color{blue}5.42} & {\color{red} 5.40}\\
     \midrule
     WRN-4-20& 5.04& 4.84 & {\color{red} 4.55} & {\color{blue}4.64}\\
     \midrule
     WRN-8-20& 4.93& 4.37 & {\color{blue}4.21} & {\color{red} 4.14}\\

    \bottomrule
  \end{tabular}
\end{table*}

\begin{table*}[htp]
  \caption{Test error of different CNNs on CIFAR-100. The data retaining rate (~$p$~) of standard dropout and DropFilter is 0.9 for all tested networks. For ScaleFilter, the retaining rate ($q$) is 0.4 except for ResNet-56 ( 0.6 ) and WRN-8-20 ( 0.2 ). The results are the average of 5 runs.}
  \label{resnet100}
  \centering
  \begin{tabular}{lcccc}
    \toprule
    Network     & without data drop(\%) & standard dropout(\%) & DropFilter(\%)  & ScaleFilter(\%) \\
    \midrule
    ResNet-56&28.07&	27.74&	{\color{blue} 25.83}&	{\color{red} 25.58}\\
    \midrule
     ResNet-110& 26.16&	25.85&	{\color{red}24.12}&	{\color{red}24.12}\\
     \midrule
     WRN-4-20& 22.90&	22.53&	{\color{red}21.24}&	{\color{blue}21.40}\\
     \midrule
     WRN-8-20& 20.99&	20.78&	{\color{blue}19.70}&	{\color{red}19.50}\\

    \bottomrule
  \end{tabular}
\end{table*}
\subsection{Different Data Retaining Rate}
The networks are regularized by DropFilter and ScaleFilter with different $p$ and $q$. Fig.~\ref{fig:diffpq:a} shows the results of standard dropout, DropFilter and ScaleFilter on CIFAR-10 with ResNet-110. The retaining rate 1 means that no data drop method is used. Standard dropout can slightly improve the networks with 0.9 retaining rate. DropFilter consistently performs better than standard dropout and significantly reduces the test error with the retaining rate of 0.9.

However, the network is very sensitive to the retaining rate of standard dropout and DropFilter, because data drop methods are applied to every convolutional layer in ResNet-110. This problem can be solved by ScaleFilter. As can be seen in Fig.~\ref{fig:diffpq:a}, ScaleFilter supplies modest and stable regularization.

The results on CIFAR-100 are shown in Fig.~\ref{fig:diffpq:b}. As can be seen, DropFilter and ScaleFilter outperform standard dropout, which is similar with the results on CIFAR-10. On CIFAR-100, the improvement of DropFilter and ScaleFilter is more   
convincing according to error bars of five runs.

\begin{table*}[ht]
  \caption{Test error of plain CNNs on CIFAR-10. The data retaining rate (~$p$~) of standard dropout and DropFilter is 0.8 and 0.95 for all tested networks. For ScaleFilter, the retaining rate ($q$) is 0.6. The results are the average of 5 runs.}
  \label{plain10}
  \centering
  \begin{tabular}{lcccc}
    \toprule
    Network     & without data drop(\%) & standard dropout(\%) & DropFilter(\%)  & ScaleFilter(\%) \\
     \midrule
     plain-4-8& 7.56&	6.86&	{\color{red}6.51}&	{\color{blue}6.60}\\
     \midrule
     plain-8-14& 5.47&	5.46&	{\color{blue}5.41}&	{\color{red}5.36}\\

    \bottomrule
  \end{tabular}
\end{table*}

\begin{table*}[ht]
  \caption{Test error of plain CNNs on CIFAR-100. The data retaining rate (~$p$~) of standard dropout and DropFilter is 0.9 and 0.95 for all tested networks. For ScaleFilter, the retaining rate ($q$) is 0.6. The results are the average of 5 runs.}
  \label{plain100}
  \centering
  \begin{tabular}{lcccc}
    \toprule
    Network     & without data drop(\%) & standard dropout(\%) & DropFilter(\%)  & ScaleFilter(\%) \\
    \midrule
     plain-4-8& 28.70&	27.83&	{\color{red}26.12}&	{\color{blue}26.54}\\
     \midrule
     plain-8-14& 24.91&	24.58&	{\color{blue}24.38}&	{\color{red}24.21}\\

    \bottomrule
  \end{tabular}
\end{table*}

\subsection{Results on ResNets and WRNs}
In this section, DropFilter is tested on ResNets and WRNs on CIFAR-10 and CIFAR-100. We apply data drop to every convolutional layer. The networks are tested using different retaining rate and the best retaining rate results are shown in Table~\ref{resnet10} and Table~\ref{resnet100}.

As shown in Table~\ref{resnet10}, DropFilter and ScaleFilter consistently outperform standard dropout for networks with different depths and widths. On CIFAR-10, WRN-4-20 with DropFilter outperforms WRN-8-20 without data drop. Note that WRN-8-20 use about four times more parameters than WRN-4-20. When DropFilter and ScaleFilter are used for WRN-8-20, its performance is further improved. For ResNet-56, DropFilter get a poor result, because it is too coarse for this network. ScaleFilter achieves a better result with a relatively low $q$ (0.6). 

On CIFAR-100, the improvement of DropFilter and ScaleFilter is more remarkable. As shown in Table~\ref{resnet100}, standard drop can only provide modest improvement. DropFilter and ScaleFilter reduce the test error by more than 1\% for all the tested networks. 

On CIFAR-100, a little bigger retaining rate can be used. WRN-8-20 uses $q=0.4$ for ScaleFilter on CIFAR-10. On CIFAR-100, $q=0.2$ is the best for this network. For ResNet-56, $p=0.9$ is a little too big for DropFilter on CIFAR-10. But on CIFAR-100, ResNet-56 performs well with the same retaining rate.

\subsection{Results on Plain Networks}
ResNets and WRNs are networks with residual connections. To demonstrate that DropFilter and ScaleFilter are useful for general convolutional networks, these two data drop methods are tested on plain networks. The plain networks are constructed by the same template as WRNs except that the residual connections are removed. Plain-14-4 represents the 14 layer network with four times more filters in each layer than ResNet template on CIFAR. Very deep plain networks suffer from the problem of vanishing/exploding gradients. Data drop methods are useless and only increase the test error for these networks. Therefore we test our methods on relatively shallow and wide plain nets(plain-8-4, plain-14-4).

As shown in Table~\ref{plain10} and Table~\ref{plain100}, for plain-8-4, DropFilter and ScaleFilter still perform better than standard dropout. Compared to plain-8-4, the margins for plain-14-4 are lower. We guess that data drop methods will deteriorate the problem of gradients. The gradients problem of deeper plain nets will be enlarged when using data drop method. Thus the margins of using data drop methods are very low for plain-14-4. That also helps to explain why data drop methods are useless for very deep plain networks.

\subsection{DropFilter with Other Methods}

In this section, more experiments are conducted demonstrate that DropFilter does not conflict with new residual networks achitecture and other training methods. The methods used in the experiments are followings: 
\begin{itemize}
\item \textbf{New Residual Architecture}.
 ResNeXt is a new residual network architecture that using group convolutions for bottleneck residual blocks~\cite{xie2016aggregated}. It introduces cardinality to residual networks and achieves better results than ResNets upon many visual recognition tasks. ResNeXt-29(16x64d) is used to test our data drop method in this paper. ResNeXt-29(16x64) uses group convolutions for the $3\times3$ convolutions. The group number is 16 and each group consists of 64 filters.
\item \textbf{Cosine Annealing for Learning Rate}.
Cosine annealing for learning rate is introduced by~\cite{loshchilov2016sgdr}. Instead of dropping the learning rate with a factor after each training stage, they drop the learning rate more smoothly by cosine annealing. Afterwards, this method is used in many researches~\cite{gastaldi2017shake,zoph2017learning}.
\item \textbf{Curriculum Dropout}.
Curriculum dropout is introduced by~\cite{Morerio2017Curriculum}. In their opinions, the co-adaptations will not occur in the beginning of training, because the networks are initialized randomly. Thus they linearly decrease the data retaining rate during training. We linearly decrease the data retaining rate of standard dropout and DropFilter as well. The retaining rate is decreased from 1.0 to 0.6.

\item \textbf{More Training Epochs}. Training the network for more epochs usually will improve the network performance on CIFAR-10\cite{loshchilov2016sgdr,gastaldi2017shake,zoph2017learning}. The training epochs are increased from 200 to 600 in our experiments.


The results on CIFAR-10 is shown in Table~\ref{resnext}. As can be seen, because of introducing too much unnecessary noise, standard dropout increases the test error. DropFilter focuses on reducing the co-adaptations between filters. It does not introducing too much useless noise and can still improve the networks. 
 
The results of competitive methods on CIFAR-10 is shown in Table~\ref{state-of-the-art}. As can be seen, improving the results on CIFAR-10 is much difficult. ResNeXt-29 (16x64d) only outperforms ResNeXt-29 (8x64d) by 0.7\%. DropFilter further improves the result nearly without introducing extra computation. Only Shake-Shake-26(2x96d) outperforms our model. But Shake-Shake-26(2x96d) needs to be trained for three times more epochs(1800 epochs).

\end{itemize}

\begin{table}[htp]
  \caption{The test error of ResNeXt-29(16x64d) with data drop regularization on CIFAR-10.}
  \label{resnext}
  \centering
  \begin{tabular}{cc}
    \toprule
    method &  test error(\%)\\
     \midrule
    without data drop&	3.42	\\
   	\midrule
    standard dropout&	3.61	\\
    \midrule
   DropFilter&	3.25	\\

    \bottomrule
  \end{tabular}
\end{table}

\begin{table*}[htp]
  \caption{Test error of our method and competitive methods on CIFAR-10.}
  \label{state-of-the-art}
  \centering
  \begin{tabular}{lcc}
    \toprule
    Method     & parameters(M) & CIFAR-10(\%) \\
    \midrule
    original-ResNet~\cite{he2016deep} & 10.2 & 7.93  \\
    \midrule
    stoc-depth-110~\cite{huang2016deep} & 1.7   & 5.23 \\
    stoc-depth-1202 & 10.2 & 4.91 \\
    \midrule
    FractalNet~\cite{larsson2016fractalnet} & 38.6 & 5.22\\
    with Dropout/Drop-path &38.6&4.60\\
    \midrule
    pre-ResNet~\cite{he2016identity} & 10.2 & 4.62   \\
    \midrule
    WRN-16-8~\cite{zagoruyko2016wide} & 36.5 & 4.00   \\ 
    WRN~(Dropout) & 36.5 & 3.80     \\
     \midrule
    ResNeXt-29~(8x64d)~\cite{xie2016aggregated} & 34.4 & 3.65 \\
    ResNeXt-29~(16x64d) & 68.1 & 3.58\\
    \midrule
    DenseNet(L = 100,k = 24)~\cite{huang2016densely} & 27.2 & 3.74 \\
    DenseNet-BC (L = 100,k = 40)~\cite{huang2017densely}&25.6 & 3.46\\  
   
    \midrule    
    NASNet-A & 3.3 & 3.41 \\
    \midrule    
    ResNeXt-29~(16x64d) DropFilter(ours) & 68.1 & {\color{blue}3.25} \\
  
    \midrule
    Shake-Shake-26 (2x96d) & 26.2 & {\color{red}2.86}\\
    \bottomrule
  \end{tabular}
\end{table*}

\subsection{Results on the Subset of ImageNet}

The methods in this paper are tested on the Subset of ImageNet. The tested network is ResNet-152. Data drop methods are used before and after the $3\times3$ convolutional layers in residual blocks. For SclaeFilter, $q=0.4$ is used. The data retaining rate for DropFilter and standard dropout is 0.9.

As shown in Table~\ref{imagenet}, standard dropout reduces the test error by about 0.3\%. DropFilter reduce the test error by 0.8\%. ScaleFilter performs best and improve the performance by about 1.1\%. DropFilter and ScaleFilter can not only work well on tiny image datasets, but also remarkably improve the results of CNNs on bigger image dataset.

\begin{table}[htp]
  \caption{The top 5 test error(\%) of ResNet152 on the subset of ImageNet. The retaining rate for standard dropout and DropFilter is 0.9. ScaleFilter uses $q =0.4$ for this network.}
  \label{imagenet}
  \centering
  \begin{tabular}{cc}
    \toprule
    method & top 5 test error(\%)\\
     \midrule
    without data drop&	11.43	\\
   	\midrule
    standard dropout&	11.12	\\
    \midrule
   DropFilter&	10.63	\\
    \midrule
   ScaleFilter & 10.34 \\

    \bottomrule
  \end{tabular}
\end{table}

\section{Discussions}

\subsection{The Data Retaining Rate}
For data drop regularization methods, the retaining rate is very important. According to our experiments, deeper and wider networks need smaller data retaining rate. If the networks are trained with curriculum in~\cite{Morerio2017Curriculum}, smaller data retaining rate can be used. For DropFilter, if 0.9 or 0.95 is still too small for the networks, we suggest using ScaleFilter rather than using DropFilter with bigger retaining rate like 0.98. DropFilter randomly drops the feature maps with the retaining rate. At the first stage of ResNet-56, there are only 16 feature maps which is too few to reveal the difference between 0.95 and 0.98 for DropFilter.

ScaleFilter can be seen as the weakened version of DropFilter. But they are a little different sometimes. Scaling the data is different from dropping the data after all. That is why DropFilter outperform ScaleFilter sometimes.

\subsection{Network Structure and Data Drop Regularization}
We propose that co-adaptations are more likely to occur between units that are equivalent in position. 
According to this, we present DropFilter and ScaleFilter to regularize convolutions. Note that this suggestion can be used for regularizing different structure networks and constructing new networks. If there is a network that should be regularized with data drop methods, we could construct new data drop method according to the co-adaptations. When we construct new network structure, we should take care of the co-adaptations that will occur. For example, when we construct multi-path networks, the networks may perform better if the paths are different from each other. That is a perspective to explain why inception~\cite{szegedy2015going,szegedy2016rethinking,szegedy2016inception} and NASNets~\cite{zoph2017learning} perform well.
\section{Conclusions}
In this paper, we propose DropFilter which is a new regularization method for convolutions. Unlike standard dropout, DropFilter only aims to reduce the co-adaptations inter filters. ScaleFilter randomly scales the outputs of filters. CNNs can be regularized by ScaleFilter when DropFilter is too coarse for the networks. Using the methods in this paper, we significantly improve the performance of different architecture CNNs on CIFAR and the subset of ImageNet. In addition, DropFilter nearly does not introduce extra computation like standard dropout.

Multi-path networks have achieved state-of-the-art results~\cite{gastaldi2017shake,zoph2017learning} on CIFAR and ImageNet with DropPath, which demonstrates that data drop regularization methods are promising for improving deep neural networks. We bridge the gaps not only between multi-path networks and single path networks but also between standard dropout and DropPath. We hope that more concerns could be paid to data drop regularization methods. We believe that this is a promising perspective to understand and improve deep neural networks.

{
\bibliographystyle{aaai}
\bibliography{egbib}
}

\end{document}